\documentclass{amia}
\usepackage{graphicx}
\usepackage[labelfont=bf]{caption}
\usepackage[superscript,nomove]{cite}
\usepackage{color}
\usepackage{comment}
\usepackage{multirow}
\usepackage{enumitem}

\begin{document}

\title{Extracting Adverse Drug Events from Clinical Notes }

\author{Darshini Mahendran and Bridget T. McInnes, Ph.D.}

\institutes{
Computer Science Department, Virginia Commonwealth University, Richmond, VA, USA
}

\maketitle

\noindent{\bf Abstract}

\textit{
Adverse drug events (ADEs) are unexpected incidents caused by the administration of a drug or medication. To identify and extract these events, we require information about not just the drug itself but attributes describing the drug (e.g., strength, dosage), the reason why the drug was initially prescribed, and any adverse reaction to the drug. This paper explores the relationship between a drug and its associated attributes using relation extraction techniques. We explore three approaches: a rule-based approach, a deep learning-based approach, and a contextualized language model-based approach. We evaluate our system on the n2c2-2018 ADE extraction dataset. Our experimental results demonstrate that the contextualized language model-based approach outperformed other models overall and obtain the state-of-the-art performance in ADE extraction with a Precision of 0.93, Recall of 0.96, and an $F_1$ score of 0.94; however, for certain relation types, the rule-based approach obtained a higher Precision and Recall than either learning approach.}

\section{Introduction}

Adverse drug events (ADE) are unexpected incidents or accidents related to the administration of a drug and its attributes. It includes overdoses, allergic reactions, drug interactions, and medication errors. ADEs account for an estimated  30\% of all hospital adverse events \cite{henry20202018}, and the cost of managing ADEs can be high because the clinical diagnosis of an ADE more often requires additional laboratory tests or procedures to investigate the cause of a patient’s symptoms. An ADE may cause a prolonged length of stay in the hospital and increase the economic burden \cite{classen1997adverse}. Some conditions can be caused by undiscovered ADEs, which can increase the costs and risks further and impact the patient economically and mentally. This can be prevented if we identify the potential reason and the information regarding the drug on time. Extracting relations from scientific publications and clinical narratives has always been challenging due to the complexity of language and domain-specific knowledge involved \cite{luo2017recurrent}. Processing the information from these clinical narratives, which records patient medication history, known allergies, reactions, and adverse events of the patient, allows a more thorough assessment of potential ADEs before they happen \cite{henry20202018}.

If we can extract all possible interactions of a drug and warn the patient when prescribing the drug, this would reduce the risks of an ADE taking place~\cite{henry20202018}. ADEs are a world-wide health-related concern, and therefore, a considerable amount of research and effort is dedicated to identifying possible ADEs in different instances~\cite{schatz2015adverse}. Information on drugs, its attributes, and associated ADEs are stored in many databases; however, we require information about not just the drug itself but attributes describing the drug (e.g., strength, dosage), the reason why the drug was initially prescribed (e.g., reason), and the relation between the drug and its attributes. 


However, manual extraction of ADEs is almost impossible~\cite{li2018extraction} given the amount of data gathered every year, therefore, there is an urgent need for automated systems for ADE extraction. These data are more often unstructured and Natural Language Processing (NLP) techniques are utilized for this significant task to extract ADEs from the unstructured text. Relation Extraction (RE) is a sub-field of NLP whose goal is to detect and classify relations between entities in a text. In this work, we explore three RE approaches for ADE extraction: a rule-based approach utilizing co-location information, a deep learning-based approach utilizing Convolutional Neural Networks (CNNs), and a contextualized language model-based approach utilizing Bidirectional Encoder Representations from Transformers (BERT). We evaluate our system on the n2c2-2018 ADE extraction dataset. Our experimental results demonstrate that the contextualized language model-based approach outperformed other models overall and obtained state-of-the-art performance in ADE extraction with a Precision of 0.93, Recall of 0.96, and an $F_1$ score of 0.94; however, the rule-based system obtained a higher Precision and Recall for certain relation types.

The remainder of this paper is structured as follows. First, we discuss the previous works done in this area of research. Second, we describe the dataset we use to evaluate our system. Third, we describe our three approaches. Fourth, we present and analyze the results. Fifth, we conduct a comparison between our approaches and previous work. Finally, we present the conclusions we derive from this work and what we plan to do in the future. 

\section{Related Work}

Adverse drug event (ADE) extraction is gaining attention recently among the clinical NLP community. Many approaches have been explored and can be divided into four paradigms: 1) rule-based, 2) machine learning-based 3) deep learning-based, and 4) contextualized language model-based approaches. 


{\it Rule-based approaches.} These systems use specified rules and patterns to extract the information from texts. Li, et al.~\cite{li2015end}, used a rule-based method to link drug names with their attributes. They used a string-based regular expression matching to match the drug names to a prescription list and then used the co-location information and RxNorm dictionary to determine whether they matched.\footnote{https://www.nlm.nih.gov/research/umls/rxnorm/index.html}. 

{\it Machine learning-based approaches.} Traditional supervised machine learning systems utilize large amounts of the annotated corpus for training.  Previous works have utilized learning algorithms such as Support Vector Machines (SVMs)~\cite{miller2019extracting} and Random Forests (RFs)~\cite{henry20202018}. Miller, et al.~\cite{miller2019extracting} used an SVM-based system and a neural system obtain for RE based on previous work on extracting temporal narrative container relations from sentences~\cite{dligach2014discovering}. Yang, et al.~\cite{yang2020identifying} first applied heuristic rules to generate candidate pairs and then applied ML models to classify the relations. They divided the relations into different groups according to their cross-distance - defined as the number of sentence boundaries between the two entities and developed multiple classifiers to classify relations according to their cross-distance.

{\it Deep learning-based approaches.} These systems utilize multi-layer neural networks typically with featureless embedding representations such as word embeddings~\cite{mikolov2013distributed}. Recent work has explored using variations of Recurrent Neural Network (RNN) architectures. Xu, et al.~\cite{xu2017uth_ccb} proposed a cascaded sequence labeling approach to recognize the entities and the relations simultaneously. Sorokin, et al.~\cite{sorokin2017context} proposed using a Long Short Term Memory (LSTM)-based encoder to jointly learn representations for all relations in a single sentence. Henry, et al.~\cite{henry20202018} summarizes the work of participants in the n2c2-2018 challenge who proposed an attention-based piecewise bidirectional (bi-) LSTM with standard features and unique candidate pair generation. Christopoulou, et al.~\cite{christopoulou2020adverse} developed separate models for intra- and inter-sentence relation extraction and combined them using an ensemble method. The intra-sentence models use biLSTMs with attention mechanisms to capture dependencies between multiple related pairs in the same sentence. For the inter-sentence relations, they used the transformer network to improve performance for longer sequences. Research on RNNs and its variants has been studied however there are few exploring CNN architectures. Therefore, in this work, we explore CNN based architectures for relation extraction.

{\it Contextualized language model-based approaches.} Pre-trained contextualized language models have been shown to increase the performance for several NLP tasks. Wei, et al.~\cite{wei2019relation} and Alimova, et al.~\cite{alimova2020multiple} applied pre-trained language models of BERT to the ADE RE task. Wei, et al.~\cite{wei2019relation} developed two BERT-based methods: Fine-Tuned BERT (FT-BERT) and Feature Combined BERT (FC-BERT) to determine relation categories for these candidate pairs. For the FT-BERT models, they represent a candidate relation pair in an input sentence by replacing the entity with its semantic type, and they added a linear classification layer on the top of The BERT model to predict the labels. For the FC-BERT, they represented the entities using the entity tags. Alimova, et al.~\cite{alimova2020multiple} proposed a machine learning model with a novel set of knowledge-based and BioSentVec embedding~\cite{chen2019biosentvec} features. For comparison, they utilized three BERT-based models: BERT\_uncased, BioBERT, and Clinical BERT. They utilized the entity texts combined with a context between them as an input for the BERT-based models.


\section{Dataset}

We evaluate our approaches on the National NLP Clinical Challenges (n2c2) 2018 Adverse Drug Event Dataset~\cite{henry20202018}. The  dataset  contains ADE mentions, drug-related attributes, and drug-related  relations from 505 patient discharge summaries drawn from the MIMIC-III database ~\cite{johnson2016mimic}. It consists of nine entity types (Drug, Strength, Route, Form, ADE, Dosage, Reason, Frequency) and eight relations between the drug entity and other non-drug entity types. Table ~\ref{tab:statistics} shows the number of relations in the training and test data. We use the gold annotated entities of this dataset for RE.\\

\begin{table}[htp]
\centering
\caption{Relation type statistics of n2c2 2018 data sets.}
\label{tab:statistics}
\scalebox{0.9}{
\begin{tabular}{|l|c|c|}
\hline
\multicolumn{3}{|c|}{\textbf{n2c2 dataset}} \\ \hline
\multicolumn{1}{|l|}{\textbf{Relation}} & \multicolumn{1}{l|}{\textbf{\# Train instances}} & \multicolumn{1}{l|}{\textbf{\# Test instances}} \\ \hline
Strength-Drug & 6702 & 4244 \\ \hline
Duration-Drug & 643 & 426  \\ \hline
Route-Drug  & 5538 & 3546 \\\hline
Form-Drug & 6654 & 4374 \\\hline
ADE-Drug & 1107 & 733 \\\hline
Dosage-Drug & 4225 & 2695  \\\hline
Reason-Drug & 5169 & 3410  \\\hline
Frequency-Drug & 6310 & 4034 \\\hline

\end{tabular}
}
\end{table}

\section{Methods}

In this work, we explore several approaches for ADE extraction: rule-based approaches, two deep learning-based approaches, and a contextualized language model-based approach. The remainder of this section describes the systems in detail. 

\subsection{Rule-based approach}
In our rule-based system, we utilize the co-location information between the drug and the non-drug entity types to determine if the non-drug entity is referring to the drug. We use a breadth-first search algorithm to find the closest occurrence of the drug on either side of the non-drug entity. For each non-drug entity, we traverse both sides until the closest occurrence of the drug is found based on the provided span values of the entities. We explore four traversal mechanisms and report the best traversal mechanism in our results: 1) traverse left-only, 2) traverse right-only, 3) traverse left-first-then-right, and 4) traverse right-first-then-left. This was conducted in two modes: 1)  limiting the traversal to only a single relation per relation type (bounded), or 2) allowing for a drug to be linked to multiple entity types with the same relation (unbounded). For example, the sentence \textit{Once her hematocrit stabilized, she was started on a heparin gtt with coumadin overlap.} contains a non-drug entity, \textit{gtt (Route)} and two drugs \textit{Heparin} and \textit{Coumadin}. The non-drug entity has a relation with the closest drug occurrence Heparin but not with Coumadin when applying the left-only traversal mechanism. 
 

\subsection{Deep learning-based approach}

Here, we describe the CNN architectures used in this work.  CNNs consist of four main layers~\cite{nguyen2015relation}: embedding, convolution, pooling, and feed-forward layers. Initially, the convolution layer which is a filter learns using the backpropagation algorithm and extracts features from the input. Then the max-pooling layer uses the position information and helps to extract the most significant features from the output of the convolution filter. Finally, the feed-forward layer uses a softmax classifier that performs classification.  CNNs take pre-trained word vectors obtained from an external resource as input. Here, we explore two word embedding types: word2vec~\cite{mikolov2013distributed} and GloVe~\cite{pennington2014glove}. We treat the RE task as a binary classification task building a separate model for each drug-entity type to determine whether a relation exists between two entities.

\begin{figure}[htp]
\centering
\includegraphics[scale=0.4]{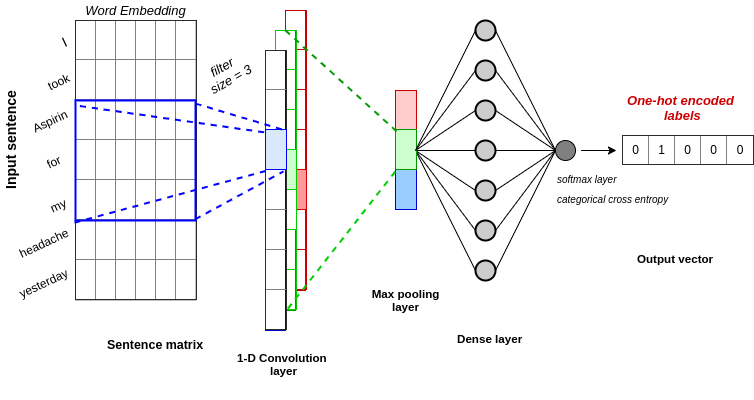}
\caption{An illustration of our model for Sentence-CNN. It explains the process of both single label and multi-label sentence CNN}
\label{fig1}
\end{figure}

\underline{{\it Sentence CNN.}}
In this architecture, for each drug-entity pair, we extract the sentence containing the relation and feed it into a CNN where each word in the sentence is represented as a vector embedding. We then apply the convolution layer to learn the local features from the embedding vectors and then the max-pooling layer to extract the most important features from the sentence. Finally, the vector is fed into a softmax (fully-connected) layer to perform the classification. The classification error is then back-propagated, and the model is re-trained until the loss is minimized. Figure \ref{fig1} shows an illustration of the Sentence CNN architecture.

\begin{figure}[htp]
\centering
\includegraphics[scale=0.4]{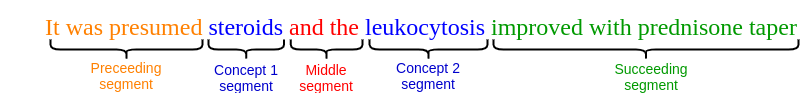}
\caption{An example of an input sentence that illustrates how the segmentation is done}
\label{fig_segments}
\end{figure}

\underline{{\it Segment-CNN.}}
In this architecture, the sentence is divided into segments and trained by separate convolutional units. First, we extract the sentence containing the relation, and we divide it into five segments: 1) preceding - tokenized words before the first concept; 2) concept 1 - tokenized words in the first concept; 3) middle - tokenized words between the two concepts 4) concept 2 - tokenized words in the second concept; and 5) succeeding - tokenized words after the second concept.
Figure~\ref{fig_segments} explains how an extracted input sentence is divided into five segments. 

\begin{figure}[htp]
\centering
\includegraphics[scale=0.35]{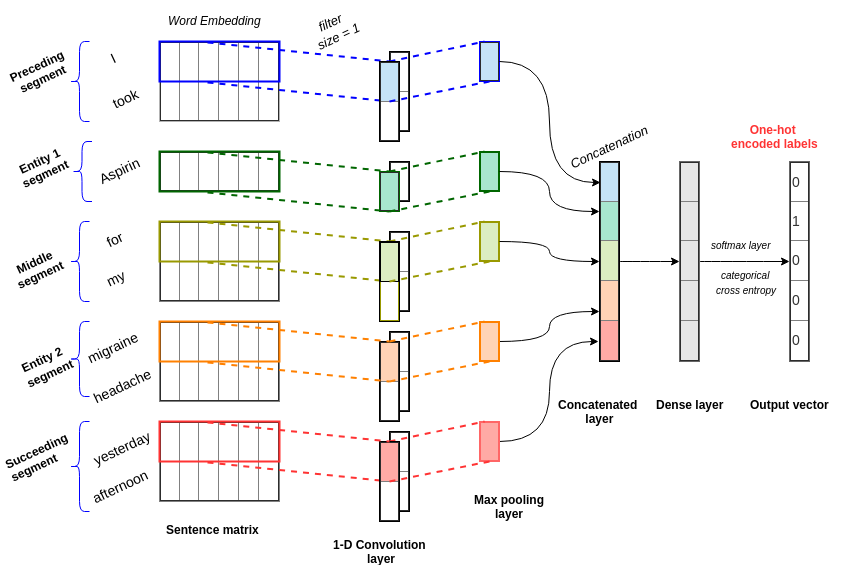}
\caption{An illustration of our model for Segment-CNN}
\label{fig2}
\end{figure}

We construct separate convolution units for each segment and concatenate them before we feed the fixed-length vector into the dense layer that performs the classification. Each convolution unit applies a sliding window that processes the segment and feeds the output to the max-pooling layer to extract essential features independent of their location. The output features of the max-pooling layer of each segment are then flattened and concatenated into a vector before feeding it into the fully connected feed-forward layer. The vector is finally fed into a softmax layer to perform the classification.  Figure \ref{fig2} shows an illustration of Segment-CNN architecture.

\subsection{Contextualized language model-based approach}

In this approach, we explore using  Bidirectional Encoder Representations from Transformers (BERT)~\cite{devlin2018bert} contextualized embeddings into a simple feed-forward neural network. We first extract the sentence containing the relation and pass it through a pre-trained BERT model. The output is then fed into a dropout layer and then into a fully-connected dense layer for classification. As with our deep learning-based approaches, we treat the RE as a binary classification task building a separate model for each drug-entity type. We explore the following BERT-based language models:
\begin{itemize}
\item {\it BERT~\cite{devlin2018bert} (-cased and -uncased)}. The original BERT models are trained on a large corpus of English data: BookCorpus (800M words) and Wikipedia (2,500M words) in a self-supervised manner (without human annotation). BERT-based models are smaller BERT models intended for environments with limited computational resources. BERT\_uncased and BERT\_cased have 2-heads, 12-layers, 768-hidden units/layer, and a total of 110 M parameters. 
\item {\it BioBERT~\cite{lee2020biobert}}. This model is initialized with the general BERT and further trained over a corpus of biomedical research articles from PubMed\footnote{https://www.ncbi.nlm.nih.gov/pubmed/} abstracts and PubMed Central\footnote{https://www.ncbi.nlm.nih.gov/pmc/} article full texts.
\item {\it Clinical BERT~\cite{alsentzer2019publicly}}.  This model is initialized with BioBERT and further fine-tuned over the Medical Information Mart for Intensive Care-III~\cite{johnson2016mimic} (MIMIC-III) clinical note corpus.
\end{itemize}

\section{Experimental design}
 
{\it Word representation}.
For our deep learning-based approaches we use Word2Vec~\cite{mikolov2013distributed} and GloVe~\cite{pennington2014glove} representations. The Word2Vec algorithm is trained over the Medical Information Mart for Intensive Care  (MIMIC-III), an openly available dataset developed by the  MIT Lab for  Computational  Physiology, comprising 2 million clinical notes from nearly 40,000 critical care patients. The gloVe is trained over Wikipedia (2014) and Gigaword 5.

{\it Text tokenization and vectorization}.
 For the rule-based and the deep learning-based approaches, we use SpaCy tokenizer\footnote{https://spacy.io/api/tokenizer} and Keras tokenizer. For the contextualized language model-based approaches, we use BertTokenizer and AutoTokenizer~\cite{alsentzer2019publicly}.
 
{\it Hyper-parameters}. 
We define our model training hyper-parameters by adjusting the batch size, learning rate, and the number of epochs. We use the batch size of 512, rmsprop optimizer with the learning rate of 0.001, and train for 10-20 epochs for our deep learning-based approach.  We used the  HuggingFaceTransformers\footnote{https://huggingface.co/transformers/} to build the BERT model for our contextualized-learning-base approach with Tensorflow 2.0. We use TFRecord to read data into a Dataset object efficiently. We use SparseCategoricalCrossentropy as the loss function and Adam as the optimizer to minimize the loss function.


\section{Evaluation criteria}

We evaluate our approaches using Precision (P), Recall (R), and $F_1$ score (F). Precision calculates out of all instances how many instances are predicted correct, and Recall calculates out of all the correct instances that should have been predicted how many instances are correctly predicted. $F_1$ score is the harmonic mean of Precision and Recall. We also report the micro and macro averages of the system performance. Micro average calculates metrics globally by counting the total true positives, false negatives, and false positives, whereas macro average calculates metrics for each label and the unweighted mean as it does not take class imbalance into account.

\section{Results and Discussion}

In this section, we describe the results of our three approaches, discuss the results across our three models, and compare previous work.

\subsection{Individual model results}

\underline{{\it Rule-based approach results.}}
Table~\ref{tab:rule_based_results} shows the Precision, Recall, and $F_1$ scores for our rule-based approach on the test set of the n2c2-2018 dataset for the top three traversal mechanisms described in our method section. Analysis of the various traversal mechanisms over all the non-drug entities showed that the {\it Left-only} traversal mechanism obtained the best results except for the entity-drug pair Duration-Drug, ADE-Drug, and Reason-Drug. Using the \textit{Left-Right (unbounded)} traversal mechanism obtained the highest $F_1$ score for these three entities. This is mainly because all other drug attributes are usually mentioned before the drug entity mentions, but the Duration, Reason, and ADE are usually mentioned after the drug mentions. Overall, This approach achieved an overall Precision of 0.88, Recall of 0.83, and $F_1$ score of 0.86. \\

\begin{table}[htp]
\centering
\caption{Results for our rule-based approaches over the n2c2-2018 test set }
\label{tab:rule_based_results}
\scalebox{0.9}{
\begin{tabular}{|l||c|c|c||c|c|c||c|c|c|}
\hline
\multirow{2}{*}{\textbf{}} & \multicolumn{3}{c||}{\textbf{Left-only}} & \multicolumn{3}{c||}{\textbf{Left-Right (unbounded)}} & \multicolumn{3}{c|}{\textbf{Left-Right (bounded)}} \\ \cline{2-10} 
 & \textbf{P} & \textbf{R} & \textbf{F} & \textbf{P} & \textbf{R} & \textbf{F} & \textbf{P} & \textbf{R} & \textbf{F} \\ \hline
Strength-Drug & 0.96 & 0.95 & \textbf{0.95} & 0.46 & 0.90 & 0.61 & 0.94 & 0.94 & 0.94 \\ \hline
Duration-Drug & 0.78 & 0.69 & \textbf{0.73} & 0.58 & 0.74 & 0.65 & 0.46 & 0.41 & 0.43 \\ \hline
Route-Drug & 0.90 & 0.89 & \textbf{0.89} & 0.45 & 0.64 & 0.53 & 0.37 & 0.36 & 0.37 \\ \hline
Form-Drug & 0.98 & 0.98 & \textbf{0.98} & 0.62 & 0.63 & 0.63 & 0.67 & 0.66 & 0.67 \\ \hline
ADE-Drug & 0.46 & 0.39 & 0.43 & 0.55 & 0.75 & \textbf{0.64} & 0.60 & 0.51 & 0.55 \\ \hline
Dosage-Drug & 0.89 & 0.89 & \textbf{0.89} & 0.61 & 0.57 & 0.59 & 0.89 & 0.88 & 0.89 \\ \hline
Reason-Drug & 0.48 & 0.35 & 0.41 & 0.61 & 0.57 & \textbf{0.59} & 0.39 & 0.28 & 0.33 \\ \hline
Frequency-Drug & 0.98 & 0.98 & \textbf{0.98} & 0.39 & 0.62 & 0.48 & 0.10 & 0.10 & 0.10 \\ \hline \hline
\textbf{System (Micro)} & 0.88 & 0.83 & \textbf{0.86} & 0.50 & 0.67 & 0.57 & 0.56 & 0.53 & 0.55 \\ \hline
\textbf{System (Macro)} & 0.85 & 0.80 & \textbf{0.83} & 0.61 & 0.70 & 0.63 & 0.58 & 0.53 & 0.55 \\ \hline
\end{tabular}
}
\end{table}

The results indicate that for most entity-drug pairs, co-location information is sufficient to identify most relations. However, the performance of the entity-drug pair ADE-Drug and Reason-Drug are lower compared to the other relation types. Our supposition for this is that the co-location information was insufficient to identify the correct ADE or Reason when multiple drugs were in the same sentence.  For example, in the sentence \textit{"Since no new infection was found this was presumed steroids and the leukocytosis improved with prednisone taper."} the non-drug entity \textit{leukocytosis} (ADE) is associated with both \textit{steroids} (Drug) and \textit{prednisone} (Drug). 

\underline{{\it Deep learning-based results.}}
Table~\ref{tab:cnn_based_results} shows the Precision (P), Recall (R) and $F_1$ scores for our Segment-CNN and Sentence CNN models over the n2c2-2018 test set. The results show that both models performed comparatively similar. In theory, we believed that Segment-CNN should have performed better because the Sentence CNN cannot differentiate the inputs when multiple drug-entity pairs are located in a sentence, but the results contradict the assumption. We believe this is because we treat this as a binary classification problem and build a separate model for each relation type. \\

\begin{table}[htp]
\centering
\caption{Results of our deep learning-based approaches over the n2c2-2018 test set}
\label{tab:cnn_based_results}
\scalebox{0.9}{
\begin{tabular}{|l||c|c|c||c|c|c|}
\hline
\multirow{2}{*}{} & \multicolumn{3}{c||}{\textbf{Segment-CNN}} & \multicolumn{3}{c||}{\textbf{Sentence CNN}} \\ \cline{2-7} 
 & \textbf{P} & \textbf{R} & \textbf{F} & \multicolumn{1}{c|}{\textbf{P}} & \multicolumn{1}{c|}{\textbf{R}} & \multicolumn{1}{c|}{\textbf{F}} \\ \hline
Strength-Drug & 0.91 & 0.88 & \textbf{0.90} & 0.90 & 0.91 & \textbf{0.90} \\ \hline
Duration-Drug & 0.39 & 0.90 & 0.55 & 0.41 & 0.90 & \textbf{0.57} \\ \hline
Route-Drug & 0.77 & 0.89 & \textbf{0.83} & 0.76 & 0.91 & \textbf{0.83} \\ \hline
Form-Drug & 0.85 & 0.95 & \textbf{0.90} & 0.85 & 0.96 & \textbf{0.90} \\ \hline
ADE-Drug & 0.32 & 0.85 & \textbf{0.46} & 0.32 & 0.85 & \textbf{0.46} \\ \hline
Dosage-Drug & 0.83 & 0.92 & \textbf{0.87} & 0.82 & 0.93 & \textbf{0.87} \\ \hline
Reason-Drug & 0.27 & 0.88 & \textbf{0.42} & 0.27 & 0.88 & 0.41 \\ \hline
Frequency-Drug & 0.56 & 0.88 & \textbf{0.69} & 0.56 & 0.88 & \textbf{0.69} \\ \hline \hline
\textbf{System (Micro)} & 0.69 & 0.90 & \textbf{0.78} & 0.68 & 0.92 & \textbf{0.78} \\ \hline
\textbf{System (Macro)} & 0.68 & 0.90 & \textbf{0.77} & 0.67 & 0.91 & \textbf{0.77} \\ \hline
\end{tabular}
}
\end{table}

Segment-CNN performed well with word2vec, whereas Sentence-CNN performed well with GloVe embeddings. We believe this is because the glove embeddings are trained over a more extensive dataset, while the word2vec MIMIC-III embeddings are trained in the same domain.

\underline{{\it Contextualized language model-based results.}}
Table~\ref{tab:BERT_results} shows the Precision (P), Recall (R) and $F_1$ scores of the four fine-tuned BERT models over the n2c2-2018 test dataset. Results show that the models obtain a similar performance overall and for each entity-drug pairs. Comparatively, BERT\_cased model performs better in some categories than the other models. \\
\begin{table}[htp]
\caption{Results of our contextualized langauge model-based approaches over the n2c2-2018 test set }
\label{tab:BERT_results}
\centering
\scalebox{0.9}{
\begin{tabular}{|l||c|c|c||c|c|c||c|c|c||c|c|c|}
\hline
\multicolumn{1}{|r|}{} & \multicolumn{3}{c||}{{\bf BERT (uncased)}} & \multicolumn{3}{c||}{{\bf BERT (cased)}} & \multicolumn{3}{c||}{{\bf BioBERT}} & \multicolumn{3}{c|}{{\bf Clinical BERT}} \\ \hline
 & \textbf{P} & \textbf{R} & \textbf{F} & \textbf{P} & \textbf{R} & \textbf{F} & \textbf{P} & \textbf{R} & \textbf{F} & \textbf{P} & \textbf{R} & \textbf{F} \\ \hline
Strength-Drug & 0.86 & 0.88 & 0.87 & 0.86 & 0.99 & \textbf{0.92} & 0.86 & 0.90 & 0.88 & 0.87 & 0.82 & 0.84 \\ \hline
Duration-Drug & 0.95 & 0.93 & 0.94 & 0.96 & 0.93 & 0.94 & 0.96 & 0.93 & \textbf{0.95} & 0.96 & 0.92 & 0.94 \\ \hline
Route-Drug & 0.92 & 0.99 & 0.95 & 0.92 & 0.97 & \textbf{0.97} & 0.92 & 0.97 & 0.94 & 0.92 & 0.95 & 0.93 \\ \hline
Form-Drug & 0.96 & 0.97 & \textbf{0.97} & 0.96 & 0.95 & 0.96 & 0.96 & 0.97 & 0.96 & 0.96 & 0.97 & \textbf{0.97} \\ \hline
ADE-Drug & 0.95 & 0.99 & \textbf{0.97} & 0.95 & 0.99 & \textbf{0.97} & 0.95 & 0.99 & \textbf{0.97} & 0.95 & 0.99 & \textbf{0.97} \\ \hline
Dosage-Drug & 0.93 & 0.96 & 0.94 & 0.93 & 0.96 & \textbf{0.95} & 0.93 & 0.96 & 0.94 & 0.93 & 0.89 & 0.91 \\ \hline
Reason-Drug & 0.96 & 0.98 & \textbf{0.97} & 0.96 & 0.98 & \textbf{0.97} & 0.96 & 0.99 & \textbf{0.97} & 0.96 & 0.99 & \textbf{0.97} \\ \hline
Frequency-Drug & 0.93 & 0.96 & \textbf{0.94} & 0.93 & 0.92 & 0.93 & 0.93 & 0.95 & \textbf{0.94} & 0.93 & 0.95 & \textbf{0.94} \\ \hline \hline
\textbf{System (Micro)} & 0.93 & 0.96 & \textbf{0.94} & \multicolumn{1}{l|}{0.93} & \multicolumn{1}{l|}{0.96} & \multicolumn{1}{l|}{\textbf{0.94}} & 0.93 & 0.95 & \textbf{0.94} & 0.93 & 0.96 & \textbf{0.94} \\ \hline
\textbf{System (Macro)} & 0.92 & 0.95 & \textbf{0.93} & \multicolumn{1}{l|}{0.92} & \multicolumn{1}{l|}{0.96} & \multicolumn{1}{l|}{\textbf{0.93}} & 0.92 & 0.95 & \textbf{0.93} & 0.92 & 0.95 & \textbf{0.93} \\ \hline
\end{tabular}
}
\end{table}
\begin{figure}[h!]
\centering
\includegraphics[scale=0.5]{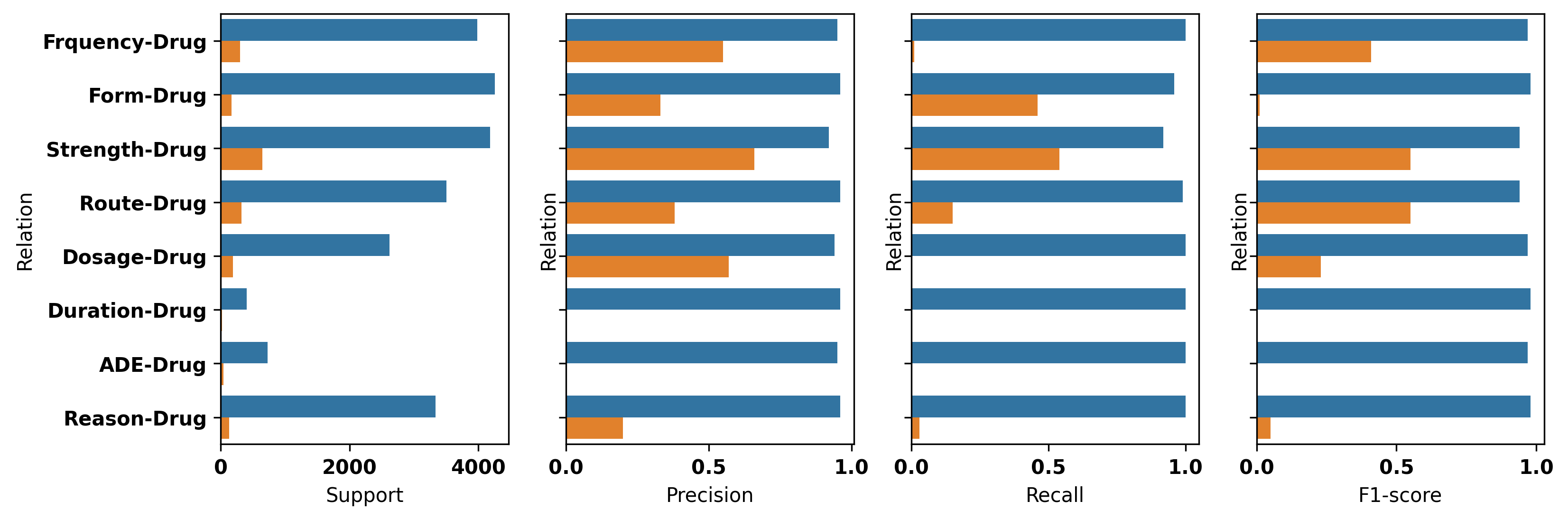}
\caption{Error analysis of each relation type during the binary classification using BERT (uncased) model. The \textit{Blue} and \textit{Brown} bars represent the positive and negative classes respectively. }
\label{fig3}
\end{figure}

\subsection{Negation Analysis}

In this work, we performed a binary classification for each class: 1) Positive class - there is a relation between the drug and the entity, 2) Negative class - there is no relation between the drug and the entity (no-relation). Figure~\ref{fig3} shows the breakdown of the performance of each class when the binary classification is performed using the BERT\_uncased model. We report Support, Precision, Recall, and $F_1$ score for each class, and \textit{Blue} and \textit{Brown} bars represent the positive and negative classes, respectively. The support shows the number of actual occurrences of the classes. We can see the Precision, Recall, and $F_1$ score of the positive classes are way higher than the negative classes. We believe this explains the higher performance of the BERT models. The performance of the negative (no-relation) classes is low due to the data imbalance of the classes, as shown in the support. Positive classes are significantly larger than the negative classes, and due to this, the poor performance of the negative classes did not affect the performance of the positive class.

\subsection{Comparison across models}

  Table~\ref{tab:comp_results} shows the Precision (P), Recall (R), and $F_1$ score for the best results of each of our three approaches: 1) rule-based approach using left-only traversal mechanism; 2) deep learning approach using Segment-CNN, and 3) contextualized language model-based approach using BioBERT. Comparing the rule-based approach with our deep learning-based approach shows that the rule-based approach obtained an overall higher Precision, Recall, and $F_1$ score except for the classes ADE-Drug and Reason-Drug. BERT-based models outperform the other two approaches except for the Strength-Drug, Frequency-Drug, and Form-Drug pairs. The overall Precision and Recall are higher, especially for the entity-drug pairs that performed poorly with the other approaches (ADE-Drug, Reason-Drug, and Duration-Drug). Using pre-trained language representations to fine-tune models is advantageous as they use minimal task-specific parameters and are trained on the downstream tasks by simply fine-tuning all the pre-trained parameters. 

\begin{table}[htp]
\centering
\caption{Comparison across our approaches over the n2c2-2018 test set }
\label{tab:comp_results}
\scalebox{0.9}{
\begin{tabular}{|l|c|c||c|c|c||c|c|c||c|c|c|}
\hline
\multirow{2}{*}{\textbf{}} & {\bf Train} & {\bf Test} & \multicolumn{3}{c||}{\textbf{Rule-based}} & \multicolumn{3}{c||}{\textbf{Segment-CNN}} & \multicolumn{3}{c|}{\textbf{BioBERT}} \\ \cline{2-12} 
& {\bf \#} & {\bf \#} & \textbf{P} & \textbf{R} & \textbf{F} & \textbf{P} & \textbf{R} & \textbf{F} & \textbf{P} & \textbf{R} & \textbf{F} \\ \hline
Strength-Drug  & 6702 & 4244 & 0.96 & 0.95 & {\bf 0.95} & 0.91 & 0.88 & 0.90 & 0.86 & 0.90 & 0.88       \\ \hline
Duration-Drug  & 643  & 426  & 0.78 & 0.69 & 0.73       & 0.39 & 0.90 & 0.55 & 0.96 & 0.93 & {\bf 0.95} \\ \hline
Route-Drug     & 5538 & 3546 & 0.90 & 0.89 & 0.89       & 0.77 & 0.89 & 0.83 & 0.92 & 0.97 & {\bf 0.94} \\ \hline
Form-Drug      & 6654 & 4373 & 0.98 & 0.98 & {\bf 0.98} & 0.85 & 0.95 & 0.90 & 0.96 & 0.97 & 0.96       \\ \hline
ADE-Drug       & 1107 & 733  & 0.46 & 0.39 & 0.43       & 0.32 & 0.85 & 0.46 & 0.95 & 0.99 & {\bf 0.97} \\ \hline
Dosage-Drug    & 4255 & 2695 & 0.89 & 0.89 & 0.89       & 0.83 & 0.92 & 0.87 & 0.93 & 0.96 & {\bf 0.94} \\ \hline
Reason-Drug    & 5169 & 3410 & 0.48 & 0.35 & 0.41       & 0.27 & 0.88 & 0.42 & 0.96 & 0.99 & {\bf 0.97} \\ \hline
Frequency-Drug & 6310 & 4034 & 0.98 & 0.98 & 0.98       & 0.56 & 0.88 & 0.69 & 0.93 & 0.95 & {\bf 0.94} \\ \hline 
\hline
\multicolumn{3}{|l||}{\textbf{System (Micro)}} & 0.88 & 0.83 & 0.86       & 0.69 & 0.90 & 0.78 & 0.93 & 0.95 & {\bf 0.94} \\ \hline
\multicolumn{3}{|l||}{\textbf{System (Macro)}} & 0.85 & 0.80 & 0.83       & 0.68 & 0.90 & 0.77 & 0.92 & 0.95 & {\bf 0.93} \\ \hline
\end{tabular}
}
\end{table}

\subsection{Comparison with previous work}

In this section, we compare our results with two previous works utilizing BERT: Wei, et al.~\cite{wei2019relation} and Alimova, et al.~\cite{alimova2020multiple} To the best of our knowledge, these are the only two works that have applied pre-trained language models of BERT on the n2c2-2018 dataset. Table \ref{tab:comparison_system} shows the overall Precision, Recall, and $F_1$ score of our fine-tuned BERT models with the reported results from the other state-of-the-art BERT-based models on the n2c2-2018 dataset. The $F_1$ score of all models of  Wei et al's and our three models is same, but the Precision of Wei, et al's models is higher whereas the Recall of our models is higher. There is a notable difference between Alimova, et al. models and ours. The $F_1$ score of all three models of Alimova, et al. are lower than ours, and we believe this is due to the difference in the representation of the inputs for the models.

\begin{table}[htp]
\centering
\caption{Overall results in comparison with previous work on the n2c2-2018 test data}
\label{tab:comparison_system}
\scalebox{0.8}{
\begin{tabular}{|l||c|c|c|c||c|c|c|c||c|c|c|}
\hline
& \multicolumn{4}{c||}{{\bf Our models}} & \multicolumn{4}{c||}{{\bf Wei, et al.~\cite{wei2019relation}}} & \multicolumn{3}{c|}{{\bf Alimova, et al.~\cite{alimova2020multiple}}} \\ \hline
& \textbf{Cased} & \textbf{Uncased} & \multicolumn{1}{l|}{\textbf{Bio}} & \textbf{Clinical} & \textbf{Cased} & \textbf{Uncased} & \textbf{Bio} & \multicolumn{1}{l|}{\textbf{Clinical}} & \textbf{Uncased} & \textbf{Bio} & \textbf{Clinical} \\ \hline
Precision & 0.93 & 0.93 & 0.93 & 0.93 & \textbf{0.98} & \textbf{0.98} & \textbf{0.98} & \textbf{0.98} & \textbf{-} & \textbf{-} & \textbf{-} \\ \hline
Recall & \textbf{0.96} & \textbf{0.96} & 0.95 & 0.93 & 0.90 & 0.90 & 0.90 & 0.90 & - & - & - \\ \hline
$F_1$ score & \textbf{0.94} & \textbf{0.94} & \textbf{0.94} & 0.93 & \textbf{0.94} & \textbf{0.94} & \textbf{0.94} & \textbf{0.94} & 0.56 & 0.75 & 0.75 \\ \hline
\end{tabular}
}
\end{table}

\begin{table}[htp]
\centering
\caption{Comparison of $F_1$ score with previous work over each class of the n2c2-2018 dataset.}
\label{tab:comparison}
\scalebox{0.8}{
\begin{tabular}{|l||c|c|c|c||c|c|c|c||c|c|c|}
\hline
& \multicolumn{4}{c||}{{\bf Our models}} & \multicolumn{4}{c||}{{\bf Wei, et al.~\cite{wei2019relation}}} & \multicolumn{3}{c|}{{\bf Alimova, et al.~\cite{alimova2020multiple}}} \\ \hline
 & \textbf{Cased} & \textbf{Uncased} & \multicolumn{1}{l|}{\textbf{Bio}} & \textbf{Clinical} & \multicolumn{1}{c|}{\textbf{Cased}} & \multicolumn{1}{c|}{\textbf{Uncased}} & \multicolumn{1}{c|}{\textbf{Bio}} & \textbf{Clinical} & \multicolumn{1}{c|}{\textbf{Uncased}} & \multicolumn{1}{c|}{\textbf{Bio}} & \multicolumn{1}{c|}{\textbf{Clinical}} \\ \hline
Strength-Drug & 0.87 & 0.87 & \multicolumn{1}{l|}{0.88} & 0.84 & 0.98 & \textbf{0.99} & 0.98 & \textbf{0.99} & 0.58 & 0.68 & 0.68 \\ \hline
Duration-Drug & \textbf{0.94} & \textbf{0.94} & \textbf{0.94} & \textbf{0.94} & 0.88 & 0.89 & 0.88 & 0.89 & 0.41 & 0.66 & 0.65 \\ \hline
Route-Drug & 0.95 & 0.95 & 0.94 & 0.93 & \textbf{0.97} & \textbf{0.97} & \textbf{0.97} & \textbf{0.97} & 0.63 & 0.74 & 0.74 \\ \hline
Form-Drug & 0.97 & 0.97 & 0.96 & 0.97 & 0.97 & \textbf{0.98} & \textbf{0.98} & \textbf{0.98} & 0.62 & 0.81 & 0.81 \\ \hline
ADE-Drug & \textbf{0.97} & \textbf{0.97} & \textbf{0.97} & \textbf{0.97} & 0.80 & 0.80 & 0.81 & 0.81 & 0.10 & 0.62 & 0.62 \\ \hline
Dosage-Drug & 0.94 & 0.94 & \multicolumn{1}{l|}{0.94} & 0.91 & \textbf{0.97} & \textbf{0.97} & \textbf{0.97} & \textbf{0.97} & 0.67 & 0.82 & 0.82 \\ \hline
Reason-Drug & \textbf{0.97} & \textbf{0.97} & \textbf{0.97} & \textbf{0.97} & 0.76 & 0.76 & 0.76 & 0.77 & 0.22 & 0.73 & 0.73 \\ \hline
Frequency-Drug & 0.94 & 0.94 & 0.94 & 0.94 & \textbf{0.96} & \textbf{0.96} & \textbf{0.96} & \textbf{0.96} & 0.53 & 0.79 & 0.78 \\ \hline
\end{tabular}
}
\end{table}

Table \ref{tab:comparison} shows a comparison of the $F_1$ score of our models the results reported by Wei, et al.~\cite{wei2019relation}'s and Alimova, et al.~\cite{alimova2020multiple} for each class of the dataset. The results show that Alimova, et al.'s models perform lower. However, when comparing the results with Wei, et al., we found the results are complementary; classes that did not perform well with Wei, et al.'s models performed well with our models. Specifically, the classes \textit{Reason-Drug, ADE-Drug, and Duration-Drug} obtained a higher Precision, Recall, and $F-1$ score than Wei, et al.'s. Meanwhile, the Precision, Recall, and $F-1$ score of the class \textit{Strength-Drug} are higher in Wei, et al. 
We believe this is due to three differences between our systems: 1) Wei, et al. represent an entity-drug pair in an input sentence using the semantic type of an entity to replace the entity itself, whereas we do no such replacement; 2) they perform a multi-class classification, whereas we perform binary classification creating a separate model for each entity; and 3) Wei, et al. Clinical BERT representations were fine-tuned with MIMIC-III over BERT (cased), whereas our representations were fine-tuned over BioBERT.  

Table~\ref{tab:competition} shows the comparison of the top five team results reported by the n2c2-2018 challenge participants~\cite{henry20202018} and our best models. The UTH system developed a joint learning biLSTM+CRF based architecture to identify the entities and relations together; the VA team used a complex traditional machine learning approach that utilized random forests; NaCT proposed a deep learning-based ensemble method; and MDQ used a biLSTM with an attention mechanism. The results show that our BERT based system obtained a higher recall across all of the systems but a lower precision across the first four teams. 

\begin{table}[htp]
\centering
\caption{Our best results in comparison with the top 5 results of the n2c2-2018 competition}
\label{tab:competition}
\scalebox{0.8}{
\begin{tabular}{|l|l|l|l|l|l|l|l|l|}
\hline
 & \multicolumn{5}{c}{\textbf{n2c2-2018 Teams}} & \multicolumn{3}{|c|}{\textbf{Our systems}}\\
 \hline
 & \textbf{UTH} & \textbf{VA} & \textbf{NaCT} & \textbf{UFL} & \textbf{MDQ} & \textbf{BERT} & \textbf{Segment-CNN} & \textbf{Rule} \\ \hline
Precision  & \textbf{0.96}  & 0.95 & 0.94 & 0.95 & 0.93 & 0.93          & 0.69 & 0.88\\ \hline
Recall      & 0.95          & 0.94 & 0.94 & 0.92 & 0.94 & \textbf{0.96} & 0.90 & 0.83\\ \hline
$F_1$ score & \textbf{0.96} & 0.94 & 0.94 & 0.94 & 0.94 & 0.94          & 0.78 & 0.86\\ \hline
\end{tabular}
}
\end{table}

\section{Conclusions and Future work}

In this work, we have investigated diverse approaches to identify relations between medication information and adverse drug events from clinical notes.  We explored a rule-based, deep learning-based, and contextualized language model-based approaches. We evaluated our approaches on the n2c2-2018 dataset and found overall the contextualized language model-based approach using BioBERT outperformed the other approaches. However, our results also showed that the rule-based approach which uses co-location information was sufficient to identify relations between entities whose positions with respect to each other were consistent throughout the text (e.g. Strength-Drug, Form-Drug and Frequency-Drug). 
In our contextualized language model-based approaches, we represent a drug-entity pair by the entire sentence, but this may not advisable as we can have multiple drug-entity pairs within a sentence. Therefore in the future, we plan to investigate effective ways of unique representations of a drug-entity pair that can capture the positional information of both the drug and entity. In this work, we developed a separate model for each class of the dataset and performed binary classification separately. In the future, we plan to investigate expanding the model to perform multi-class classification for different datasets.

\makeatletter
\renewcommand{\@biblabel}[1]{\hfill #1.}
\makeatother

\bibliographystyle{vancouver}
\bibliography{refs}

\end{document}